\documentclass{article} 
\usepackage{iclr2027_conference,times}

\usepackage{amsmath,amsfonts,bm}

\def\eqref#1{equation~\ref{#1}}

\def\1{\bm{1}}

\DeclareMathAlphabet{\mathsfit}{\encodingdefault}{\sfdefault}{m}{sl}
\SetMathAlphabet{\mathsfit}{bold}{\encodingdefault}{\sfdefault}{bx}{n}

\usepackage{hyperref}
\usepackage{url}
\usepackage{amsmath}
\usepackage{amssymb}
\usepackage{graphicx}
\usepackage{booktabs}
\usepackage{algorithm}
\usepackage{algorithmic}
\usepackage{xcolor}

\title{Meet Students Where They Are: \\Student-Prefix Teacher Adaptation for \\On-Policy Distillation}

\author{\textbf{Xue Zhang\textsuperscript{1,2}\thanks{This work was done during the internship at Weixin AI, Tencent Inc, China.}},
\textbf{Songming Zhang\textsuperscript{3}},
\textbf{Yunlong Liang\textsuperscript{3}},
\textbf{Fandong Meng\textsuperscript{3}},\\
\textbf{Yufeng Chen\textsuperscript{1,2}\thanks{Yufeng Chen is the corresponding author.}},
\textbf{Jinan Xu\textsuperscript{1,2}}\\
\textsuperscript{1}Key Laboratory of Big Data \& Artificial Intelligence in Transportation, Beijing Jiaotong University\\
\textsuperscript{2}School of Computer Science and Technology, Beijing Jiaotong University, Beijing, China\\
\textsuperscript{3}Weixin AI, Tencent, China\\
\texttt{\{zhang\_xue\}@bjtu.edu.cn}}

\newcommand{\spta}{\textnormal{\textsc{SPTA}}}
\hypersetup{hypertexnames=false}
\renewcommand{\eqref}[1]{\textup{(\ref{#1})}}

\iclrfinalcopy
\begin{document}
\maketitle

\begin{abstract}

On-policy distillation (OPD) mitigates the student's exposure bias by optimizing on student-generated trajectories.
However, student-generated prefixes may contain reasoning errors or follow patterns unfamiliar to the teacher, undermining the reliability of its token-level supervision.
This degradation becomes more pronounced as prefixes grow longer.
To address this teacher-side distribution mismatch, we propose \textbf{Student-Prefix Teacher Adaptation} (SPTA), which adapts the teacher to student-generated prefixes before distillation to improve the reliability of its token-level supervision.
Specifically, before OPD starts, we collect student rollout prefixes with random lengths and continually train the teacher to continue from the prefixes using reinforcement learning with verifiable rewards (RLVR).
By treating each prefix as context and rewarding correct final answers, the teacher learns to recover from student-induced states rather than imitate the student's reasoning.
Then, the adapted teacher is used for standard OPD, with no changes to the distillation objective.
Experiments on mathematical reasoning benchmarks show that our SPTA improves average student performance over standard OPD and OPD with task-level RLVR-adapted teachers.
These results highlight the importance of adapting teachers not only to solve problems independently, but also to provide reliable guidance on the states their students actually visit.

\end{abstract}

\section{Introduction}
\label{sec:intro}

Knowledge distillation aims to transfer the predictive behavior of a larger teacher model to a smaller student model \citep{hinton2015distilling}.
For autoregressive language models, however, training on teacher-generated sequences \citep{seqkd} creates a distributional mismatch between the prefixes observed during training and those produced by the student at inference time, a phenomenon known as exposure bias \citep{bengio2015scheduledsampling,ranzato2015sequence}.
On-policy distillation (OPD) closes this mismatch by sampling trajectories from the student itself and collecting token-level teacher supervision over these on-policy samples \citep{gkd,minillm}. 

\begin{figure}[t]
\centering
\includegraphics[width=0.95\linewidth]{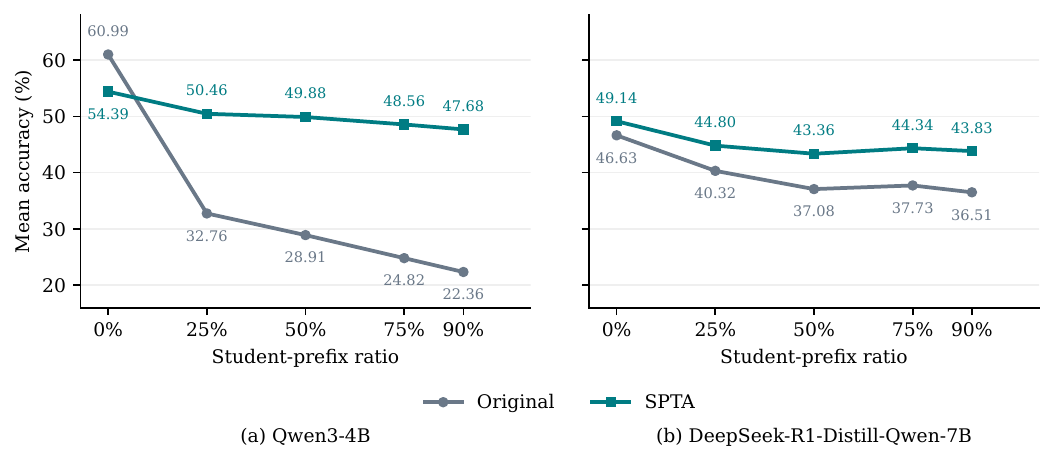}
\caption{\textbf{Student-prefix adaptation improves teacher recovery.}
Mean accuracy (avg@$16$ across five benchmarks) versus the fraction of a student rollout retained as context.
Each panel compares the original teacher with its \spta{}-Iter1 version on identical sampled student prefixes; 0\% denotes clean-prompt generation.
Recovery improves at every evaluated nonzero ratio, even though Qwen3-4B's clean-prompt accuracy decreases.}
\label{fig:motivating}
\end{figure}

However, the reliability of this token-level supervision depends critically on the teacher's conditional distribution $\pi_{T}(\cdot\mid x,y^S_{<t})$, which is conditioned on student-generated prefixes $y^S_{<t}$.
These prefixes may contain flawed reasoning paths, unfamiliar phrasing, or partial intermediate states unseen during the teacher's task-oriented training, which introduces a teacher-side distribution mismatch.
The resulting distribution mismatch degrades the teacher's supervision quality, and the issue becomes more pronounced as prefixes grow longer \citep{revisitingopd,pruneopd,xie2026position}.
As shown in Figure~\ref{fig:motivating}, Qwen3-4B's accuracy drops from 60.99\% when solving problems from scratch to 28.91\% when given prefixes covering 50\% of student rollouts, and further to 22.36\% at a prefix ratio of 90\%.
DeepSeek-R1-Distill-Qwen-7B exhibits a similar decline, with accuracy dropping from 46.63\% without a student prefix to 37.08\% at prefix ratios of 50\%.
These results suggest that a teacher's strong standalone performance does not guarantee reliable guidance on student-generated prefixes.

\begin{figure}[t]
\centering
\includegraphics[width=0.95\linewidth]{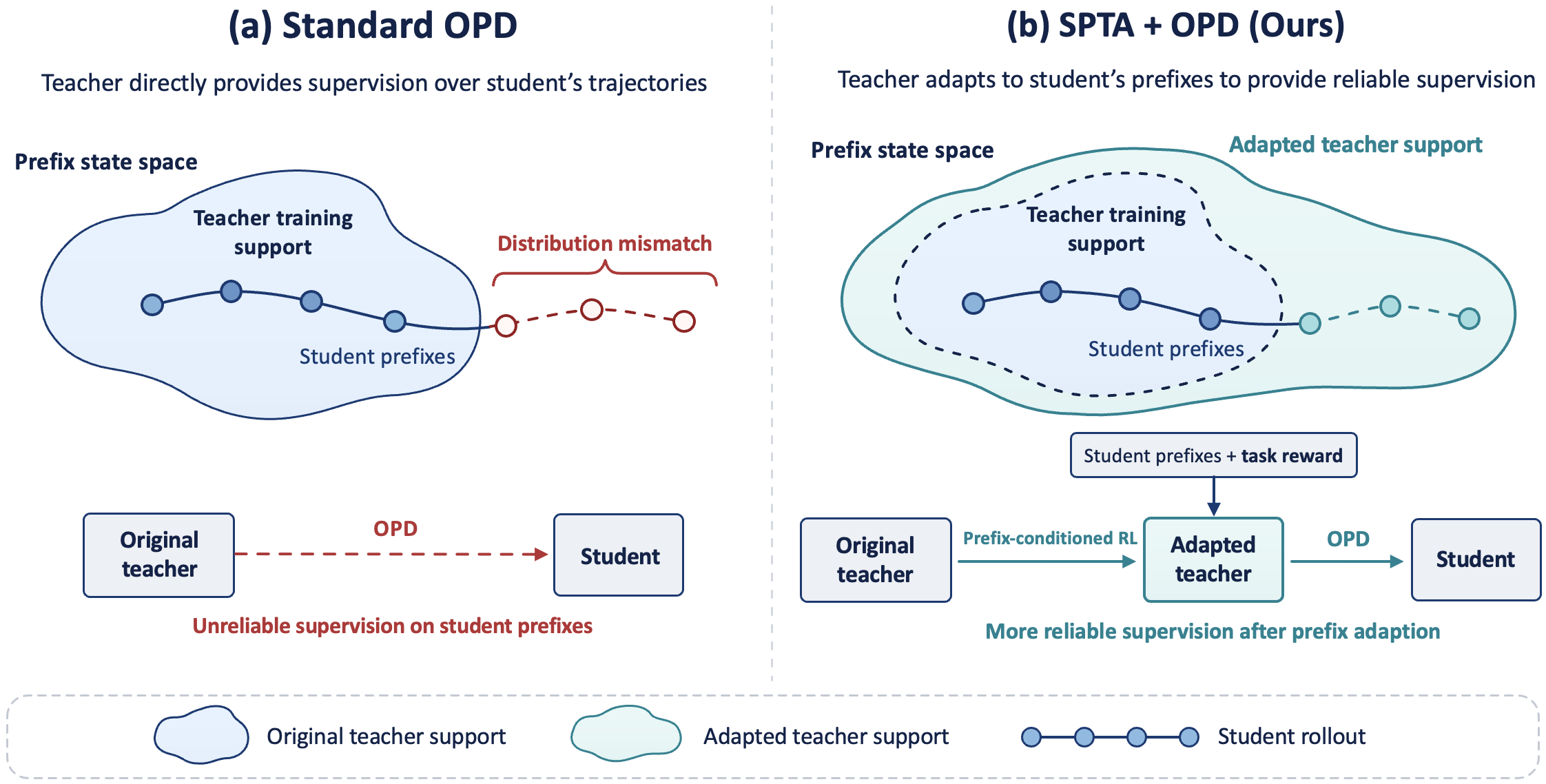}
\caption{\textbf{Standard OPD and Student-Prefix Teacher Adaptation (\spta{}).}
\textbf{(a)} In standard OPD, student rollouts may enter regions unfamiliar to the teacher, reducing supervision reliability.
\textbf{(b)} \spta{} first adapts the teacher through prefix-conditioned RLVR, using student prefixes as context and final-answer correctness as reward, then freezes it for standard OPD.
Shaded regions indicate prefixes familiar to the teacher before and after adaptation.}
\label{fig:pipeline}
\end{figure}

Existing approaches mitigate unreliable teacher guidance by adjusting where and how teacher supervision is used, for example through rollout truncation \citep{earlystoppingopd,pruneopd}, prefix selection \citep{prefixguidedopd}, or token-level signal selection and reweighting \citep{xie2026position,xu2026tip,liu2026teacher}.
On the teacher side, \citet{manyfaces} strengthen the teacher with RLVR before distillation, without directly addressing the distribution mismatch induced by student-generated prefixes. 
This motivates \textbf{Student-Prefix Teacher Adaptation} (\spta{}), a simple and straightforward approach that adapts the teacher to student-generated prefixes before distillation (Figure~\ref{fig:pipeline}(b)).

\spta{} consists of two stages.
First, we collect student rollouts before OPD and truncate them at randomly sampled positions to obtain prefixes of varying lengths.
We then adapt the teacher using RLVR to continue from these prefixes, with final-answer correctness as the reward.
Through this pre-OPD adaptation, the teacher learns to generate correct solutions conditioned on student-generated prefixes and can thus provide more reliable supervision on these prefixes.
Second, we freeze the adapted teacher and use it to provide token-level supervision on student rollouts, keeping the OPD objective unchanged.

Across five mathematical reasoning benchmarks, a single adaptation round improves mean student accuracy in all three teacher--student pairings we evaluate.
In the two pairings that include a task-level RLVR control, prefix-conditioned adaptation yields the stronger student under matched teacher optimization schedules, which indicates that the gain comes from \emph{where} the teacher is optimized rather than from how much extra training it receives.
The effect is largest where it matters most: in one pairing, OPD with the original teacher improves the student by a mere 0.62 points over no distillation at all, and prefix adaptation is what converts this stalled pipeline into a 3.65-point gain.
A second adaptation round pushes the Qwen3-4B student further still, at the cost of one more round of training.
Taken together, these results suggest that teachers should be prepared not only to solve problems on their own, but also to give reliable guidance on the states their students actually visit.

In summary, the major contributions of this paper are as follows:
\begin{enumerate}
  \item We show that standalone problem-solving performance is insufficient to assess a teacher's effectiveness on student-generated prefixes, highlighting prefix-conditioned performance as an important consideration for teacher adaptation in OPD.

  \item We introduce \spta{}, which directly optimizes the teacher on student-generated prefixes using RLVR and then uses the adapted teacher for standard OPD, keeping the distillation objective unchanged.

  \item Across five mathematical reasoning benchmarks and three teacher--student pairings, \spta{} significantly improves average student accuracy over standard OPD. Comparisons under matched teacher optimization schedules further show that prefix-conditioned adaptation yields stronger students than task-level adaptation.
\end{enumerate}

\section{Related Work}
\label{sec:related}

\paragraph{On-policy distillation and reliable supervision.}
Classical knowledge distillation matches teacher predictions \citep{hinton2015distilling}, while sequence-level distillation trains on teacher-generated outputs \citep{seqkd}.
To reduce the resulting training--inference mismatch, GKD learns from student-generated sequences with flexible divergence objectives \citep{gkd}, and MiniLLM uses on-policy reverse-KL optimization \citep{minillm}.
A recent survey \citep{opdsurvey} organizes this literature around objectives \citep{gkd,minillm,kodistillm,jung-etal-2025-todi,wu-etal-2025-rethinking,zhang-etal-2025-aligndistil}, supervision sources \citep{zhao2026opsd}, cross-tokenizer scenarios \citep{zhang-etal-2024-dual,zhang2025dskdv2,sun2026simct}, and systems \citep{verl,slime_github,zhang-etal-2026-kdflow}; yet matching the student's state distribution does not guarantee reliable teacher feedback.
\citet{revisitingopd} identify unreliable guidance on student prefixes alongside token-level supervision imbalance and tokenizer mismatch, motivating teacher top-$K$ local support matching and sampling safeguards.
To counter supervision fidelity decay, Lookahead Group Reward scores candidate tokens by the next-step teacher confidence they induce \citep{supervisiondecay}.
Other approaches modify the supervision budget: Early Stopping Rollout retains only initial response segments \citep{earlystoppingopd}; Prune-OPD down-weights unreliable rewards and dynamically truncates drifted trajectories \citep{pruneopd}; and PG-OPD allocates long rollouts using early-prefix teacher--student top-$k$ overlap \citep{prefixguidedopd}.
In multi-turn tasks, ReOPD replays teacher prefixes with a step-decaying schedule favoring earlier, lower-shift histories \citep{reopd}.
These methods improve the signal or select where to apply it; \spta{} instead adapts the teacher before standard OPD, without changing the distillation objective or filtering its rollouts.

\paragraph{Student-aware teacher adaptation.}
A complementary direction adapts the teacher for transfer rather than judging it solely by standalone performance.
PromptKD tunes soft prompts for a frozen teacher using student guidance \citep{promptkd}.
\citet{manyfaces} analyze student-prefix mismatch and optimization instability, and show that RLVR-adapted teachers can mitigate OPD failures, motivating our task-level teacher-RL control.
Outcome-based optimization is itself a proven route to stronger reasoning \citep{deepseekr1,justrl}, yet enriching teacher supervision is not invariably beneficial: privileged-context self-distillation can degrade thinking models by perturbing signals at reasoning forks and self-correction steps \citep{rethinkingselfdistill}.
\spta{} targets a different adaptation task: it updates teacher parameters using verifiable final-answer rewards on continuations of student-generated prefixes, then freezes the teacher for OPD.
The reward encourages successful completion rather than agreement with the student, so erroneous reasoning can be corrected; reference answers enter only the reward, never the teacher's context during distillation.

\section{Teacher-Side Distribution Shift in OPD}
\label{sec:shift}

\subsection{Setup and Notation}

Let $\pi_S(\cdot;\theta)$ denote the student model parameterized by $\theta$, $\pi_{T_0}(\cdot;\phi_0)$ the original teacher model parameterized by $\phi_0$, and $\mathcal{D}$ the distribution of training prompts.
For a prompt $x\sim\mathcal{D}$ and student trajectory $y^S\sim\pi_S(\cdot\mid x)$, the token-level OPD objective is
\begin{equation}
\mathcal{L}_{\mathrm{OPD}}(\theta;\phi_0)
= \mathbb{E}_{x\sim\mathcal{D},\; y^S\sim\pi_S(\cdot\mid x)}
\biggl[\sum_{t=1}^{|y^S|} D_{\mathrm{KL}}\Bigl(\pi_S(\cdot\mid x,y^S_{<t}) \parallel \pi_{T_0}(\cdot\mid x,y^S_{<t})\Bigr)\biggr],
\label{eq:opd}
\end{equation}
where $D_{\mathrm{KL}}$ measures the KL divergence between next-token predictive distributions, and the teacher $\pi_{T_0}$ remains frozen during optimization with no gradients propagated through it.
The objective thus separates \emph{which states receive supervision} from \emph{the quality of the teacher targets there}: student rollouts address the former by construction, but do nothing to optimize the latter.

We assess prefix-conditioned behavior using \textbf{recovery accuracy}.
Fix an evaluation student $\pi_S$ and a benchmark prompt distribution $\mathcal{E}$.
For a rollout $y^S$ of length $L$ and ratio $r$, let $k=\lfloor rL\rfloor$ and sample a teacher continuation $c\sim\pi_\phi(\cdot\mid x,y^S_{1:k})$.
Then
\begin{equation}
\mathrm{Rec}(\pi_\phi;r\mid\pi_S)
= \mathbb{E}_{x\sim\mathcal{E},\,y^S\sim\pi_S,\,c\sim\pi_\phi}
\bigl[\mathrm{Ver}(x,y^S_{1:k}\oplus c)\bigr],
\label{eq:rec}
\end{equation}
where $\mathrm{Ver}\in\{0,1\}$ checks the final answer against the reference and $\oplus$ denotes concatenation.
The empty prefix at $r=0$ recovers clean-prompt accuracy.
``Recovery'' denotes final-answer success after a prefix, whether or not that prefix contains an error; it is an outcome-level diagnostic and certifies neither every intermediate step nor the token-level targets OPD consumes.

\subsection{Motivating Experiment}
\label{sec:motivating}

We compare the original and \spta{}-adapted teachers after the first adaptation iteration at $r\in\{0\%,25\%,50\%,75\%,90\%\}$ on AIME24, AIME25, AMC23, HMMT24, and HMMT25.
Prefixes are sampled from Qwen3-4B-Base for the Qwen3-4B teacher and from DeepSeek-R1-Distill-Qwen-1.5B for the DeepSeek-R1-Distill-Qwen-7B teacher (configurations (1) and (2)).
Within each pairing and ratio, both teacher variants receive exactly the same sampled student prefixes.
We report avg@$16$ accuracy and its unweighted mean across the five benchmarks; Appendix~\ref{app:recovery} gives the full breakdown.

Figure~\ref{fig:motivating} shows three findings.
\textbf{O1: Original teachers lose accuracy on student prefixes.}
At $r{=}50\%$, Qwen3-4B falls from 60.99\% to 28.91\% ($-32.08$ points), while DeepSeek-R1-Distill-Qwen-7B falls from 46.63\% to 37.08\% ($-9.55$ points).
At $r{=}90\%$, their accuracies are 22.36\% and 36.51\%, respectively.
\textbf{O2: Student-prefix adaptation improves recovery across ratios.}
\spta{} raises accuracy at $r{=}50\%$ to 49.88\% and 43.36\% ($+20.97$ and $+6.28$ points).
At $r{=}90\%$, the gains are $+25.32$ and $+7.32$ points, and every evaluated nonzero ratio shows a positive mean gain in both pairings.
\textbf{O3: Recovery and clean-prompt accuracy can move differently.}
Qwen3-4B's clean accuracy decreases by 6.60 points, whereas DeepSeek-R1-Distill-Qwen-7B's increases by 2.51 points.
Thus improved recovery need not come from improved independent problem solving.
These comparisons establish the effect of adaptation on matched student prefixes; without an own-prefix control, they do not isolate prefix source from context length or continuation difficulty.

\section{Student-Prefix Teacher Adaptation}
\label{sec:method}

\subsection{Overview}

\spta{} implements the student-aware adaptation illustrated in Figure~\ref{fig:pipeline}(b) through two stages.
\textbf{Stage~I: prefix-conditioned teacher adaptation.} We collect student rollouts, truncate each at a uniformly sampled point, and run RLVR on teacher continuations from the resulting prefixes, with the RL action loss applied only to the continuation.
\textbf{Stage~II: OPD with a frozen adapted teacher.} We freeze $\pi_{\phi^\star}$ and use it to provide token-level targets on fresh student rollouts, keeping the OPD objective unchanged.
One-round comparisons initialize the student from the same $S_0$ under matched distillation settings, whereas the iterative extension of Section~\ref{sec:opd_adapted} continues from previously trained checkpoints.

\subsection{Random Student-Prefix Construction}

For each training prompt $x$, a fixed source student $\pi_S$ generates $m{=}4$ rollouts and we retain one of them uniformly at random, so each prompt contributes a single prefix; $\pi_S=S_0$ in the first iteration.
For the retained rollout $y^S=(y_1^S,\ldots,y_L^S)$, we sample
\[
r\sim\mathcal{U}(0.01,0.90),\qquad k=\lfloor rL\rfloor,\qquad
h^S=(x,y^S_{1:k}).
\]
The resulting prefix set $\mathcal{B}$ is fixed during that teacher-adaptation run.
Sampling relative positions exposes the teacher to different stages of a student's solution without requiring an error detector or a correctness filter.
It is a coverage heuristic rather than an exact match to the student's token-level occupancy distribution; we use the same range in every experiment and did not tune it.

\subsection{Prefix-Conditioned Teacher RLVR}

Starting from $\phi_0$, the teacher samples a continuation $c\sim\pi_\phi(\cdot\mid h^S)$ and receives $R(x,y^S_{1:k},c)=\mathrm{Ver}(x,y^S_{1:k}\oplus c)$.
The reward objective is
\begin{equation}
J_{\mathrm{prefix}}(\phi)
=\mathbb{E}_{h^S\sim\mathcal{B},\,c\sim\pi_\phi(\cdot\mid h^S)}
\bigl[R(x,y^S_{1:k},c)\bigr],\qquad
\max_\phi J_{\mathrm{prefix}}(\phi).
\label{eq:rlvr}
\end{equation}
Only continuation tokens contribute action log-probabilities to the RL loss: student-prefix tokens are conditioning context, not supervised teacher actions.
The reward evaluates the final answer, so the objective permits the teacher to correct or abandon an erroneous line of reasoning rather than imitate it.
It does not require identifying the first error or providing a reference continuation.
A task-level RLVR teacher uses the same optimization procedure but starts from $x$ alone; the distinction is the initial state supplied to the teacher.

\paragraph{Why an outcome reward should improve token-level targets.}
Eq.~\eqref{eq:spta} reads the student's supervision off $\pi_{\phi^\star}(\cdot\mid x,y^S_{<t})$, so what OPD needs is a teacher that places mass on tokens leading to a correct completion \emph{from that prefix}.
Eq.~\eqref{eq:rlvr} optimizes exactly this conditional: it raises the likelihood of verified continuations from student-induced states, and because a continuation is a token sequence, the update redistributes mass at every position along it rather than only at a final answer token.
Prefix-conditioned RLVR and OPD therefore act on the same family of conditionals $\{\pi_\phi(\cdot\mid x,y^S_{<t})\}$, whereas task-level RLVR improves the teacher at $t{=}0$ and relies on generalization to reach the prefixes OPD queries.
This motivates the design rather than proving it: the outcome reward does not directly supervise per-token calibration, so we test the prediction empirically in Sections~\ref{sec:main} and~\ref{sec:capability}.

\subsection{OPD with the Adapted Teacher}
\label{sec:opd_adapted}

Freeze the adapted teacher $\pi_{\phi^\star}$, initialize the student from $S_0$, and optimize
\begin{equation}
\mathcal{L}_{\mathrm{SPTA}}(\theta)
= \mathbb{E}_{x\sim\mathcal{D},\,y^S\sim\pi_\theta}\Bigl[\sum_{t=1}^{|y^S|}
D\bigl(\pi_\theta(\cdot\mid x,y^S_{<t}),\;
\pi_{\phi^\star}(\cdot\mid x,y^S_{<t})\bigr)\Bigr].
\label{eq:spta}
\end{equation}
Relative to Eq.~\eqref{eq:opd}, the only change in the distillation objective is $\phi_0\to\phi^\star$.
Thus \emph{\spta{} changes the teacher, not the distillation objective}.
This separation makes teacher adaptation compatible in principle with different OPD losses, without establishing empirical robustness across those losses.

\paragraph{Iterative adaptation.}
Since the student's prefix distribution shifts during distillation, the two stages can also be repeated: a later round collects fresh prefixes from the distilled student and warm-starts both teacher and student from their preceding checkpoints instead of resetting them to $T_0$ and $S_0$.
A single round is our default; Section~\ref{sec:iteration_results} reports one additional round, which combines refreshed prefixes with continued teacher and student training and is therefore neither a restart-from-$S_0$ experiment nor a compute-matched comparison with the single-round setting.

\section{Experimental Setup}
\label{sec:setup}

\paragraph{Teacher--student configurations.}
We evaluate three pairings with students of 4B and 1.5B parameters:
(1) \textbf{Qwen3-4B $\to$ Qwen3-4B-Base}, a same-size pairing with different post-training histories \citep{qwen3};
(2) \textbf{DeepSeek-R1-Distill-Qwen-7B $\to$ DeepSeek-R1-Distill-Qwen-1.5B}, a larger-to-smaller pairing \citep{deepseekr1}; and
(3) \textbf{JustRL-DeepSeek-1.5B $\to$ DeepSeek-R1-Distill-Qwen-1.5B}, a same-size pairing whose teacher was obtained by RL from the student's initialization \citep{justrl}.
Configurations (2) and (3) share the student but differ in teacher scale.

\paragraph{Training and evaluation.}
We conduct our experiments based on the DAPO-Math-17k \citep{dapomath} dataset.
We evaluate on AIME24, AIME25, AMC23, HMMT24, and HMMT25.
The evaluation settings are following: temperature $0.7$, top-$p$ $0.95$, and avg@$16$.
The reported Avg.\ is the unweighted mean over the five benchmarks; all scores are percentages and all gains are absolute points.
Teacher recovery uses the same benchmarks with matched student prefixes (Section~\ref{sec:motivating}); every adapted teacher in that study is from Iter1.

\paragraph{Baselines and controls.}
\textbf{OPD} uses the original, unadapted teacher.
\textbf{Tea-RL-OPD} first adapts the teacher with task-level RLVR from clean prompts, using the same RL algorithm, hyperparameters, and update count as one round of \spta{}.
\textbf{\spta{}-OPD} instead conditions teacher RLVR on student prefixes.
Tea-RL-OPD is evaluated in configurations (1) and (2); configuration (3) compares OPD with \spta{} only, since JustRL-DeepSeek-1.5B is trained using GRPO based on the student model (\textit{i.e.}, Tea-RL-OPD).
For context we also quote \textbf{Prune-OPD} \citep{pruneopd}, a published rollout-pruning method reported on the same student and benchmarks, in configurations (2) and (3).
Its numbers are taken from that paper and were not reproduced under our settings, so they share the teacher--student pairing but not the training pipeline, and we mark them accordingly in Table~\ref{tab:main}.

\paragraph{Implementation.}
All teacher--student pairings use the same training settings: GRPO \citep{deepseekmath} for teacher adaptation and policy-gradient OPD with the sampled reverse-KL \texttt{k1} signal for student training, without task rewards in the OPD loss.
Both stages use AdamW \citep{adamw} at learning rate $10^{-6}$ with constant scheduling in \texttt{verl} \citep{verl} and vLLM rollouts \citep{vllm}.
Teacher batches contain 256 prefix-conditioned inputs with eight continuations each; student batches contain 64 prompts with four rollouts each.
Appendix~\ref{app:impl} gives the two stages' distinct length limits, optimization settings, and recorded training schedules.


\begin{table}[t]
\caption{\textbf{Student accuracy after distillation} (\%, avg@$16$). Avg.\ is the unweighted benchmark mean. All distillation runs use a single adaptation/distillation round, and within each pairing they share initialization and settings. Bold marks the best result in each column, including ties; the undistilled $S_0$ row means the original student performance. $^{*}$Prune-OPD scores are quoted from \citet{pruneopd} and were not reproduced by us, so they are listed for context rather than as a controlled comparison.}
\label{tab:main}
\centering
\small
\setlength{\tabcolsep}{4pt}
\begin{tabular}{lcccccc}
\toprule
\textbf{Method} & \textbf{AIME24} & \textbf{AIME25} & \textbf{AMC23}
  & \textbf{HMMT24} & \textbf{HMMT25} & \textbf{Avg.} \\
\midrule
\multicolumn{7}{l}{\textit{(1) Qwen3-4B $\rightarrow$ Qwen3-4B-Base}} \\
Student $S_0$ (no distillation) & 12.50 & 6.04 & 38.48 & 4.58 & 0.83 & 12.49 \\
OPD & 33.54 & 26.04 & 65.51 & 16.04 & 16.46 & 31.52 \\
Tea-RL-OPD & 33.12 & 27.29 & 65.89 & \textbf{16.67} & 16.88 & 31.97 \\
\spta{}-OPD & \textbf{37.71} & \textbf{33.54} & \textbf{71.91} & 16.04 & \textbf{17.29} & \textbf{35.30} \\
\midrule
\multicolumn{7}{l}{\textit{(2) DeepSeek-R1-Distill-Qwen-7B $\rightarrow$ DeepSeek-R1-Distill-Qwen-1.5B}} \\
Student $S_0$ (no distillation) & 32.29 & 23.96 & 63.33 & 15.21 & 13.13 & 29.58 \\
OPD & 31.04 & 23.75 & 64.76 & 13.96 & 17.50 & 30.20 \\
{Prune-OPD}$^{*}$ & 33.30 & 25.80 & 65.90 & 14.60 & 15.00 & 30.92 \\
Tea-RL-OPD & 33.12 & \textbf{27.50} & \textbf{68.67} & \textbf{16.46} & 16.04 & 32.36 \\
\spta{}-OPD & \textbf{38.96} & 25.21 & 68.00 & 16.25 & \textbf{17.71} & \textbf{33.23} \\
\midrule
\multicolumn{7}{l}{\textit{(3) JustRL-DeepSeek-1.5B $\rightarrow$ DeepSeek-R1-Distill-Qwen-1.5B}} \\
Student $S_0$ (no distillation) & 32.29 & 23.96 & 63.33 & 15.21 & 13.13 & 29.58 \\
OPD (Tea-RL-OPD) & 47.92 & \textbf{35.42} & \textbf{85.47} & 18.75 & 22.29 & 41.97 \\
{Prune-OPD}$^{*}$ & 45.20 & 33.50 & 78.40 & 22.70 & 18.80 & 39.72 \\
\spta{}-OPD & \textbf{52.50} & 34.58 & 80.20 & \textbf{23.75} & \textbf{22.92} & \textbf{42.79} \\
\bottomrule
\end{tabular}
\end{table}

\section{Main Results}
\label{sec:main}

Table~\ref{tab:main} compares the distilled students. All rows use a single adaptation/distillation round, so the primary comparison is \spta{}-OPD against OPD and, where evaluated, Tea-RL-OPD; the effect of a second round is deferred to Section~\ref{sec:iteration_results}.

\paragraph{Prefix adaptation improves mean student accuracy across pairings.}
For Qwen3-4B $\to$ Qwen3-4B-Base, \spta{} raises the average from 31.52 to 35.30 ($+3.78$ points).
With DeepSeek-R1-Distill-Qwen-7B as teacher, it raises the average from 30.20 to 33.23 ($+3.03$ points).
With JustRL-DeepSeek-1.5B, it raises the average from 41.97 to 42.79 ($+0.82$ points).
The gains span both same-size and larger-to-smaller distillation.

\paragraph{With the original teacher, OPD can be close to a no-op.}
The undistilled student $S_0$ changes how the OPD baseline should be read.
In configuration (1) the student is a base model with ample headroom, and OPD already gains 19.03 points over $S_0$ (12.49 to 31.52); \spta{} adds 3.78 on top.
In configuration (2), by contrast, OPD improves over $S_0$ by only 0.62 points (29.58 to 30.20) and is \emph{below} $S_0$ on three of the five benchmarks, so distillation from the original DeepSeek-R1-Distill-Qwen-7B teacher barely moves the student at all.
\spta{} turns this into a 3.65-point gain and improves on $S_0$ on all five benchmarks.
Prefix adaptation is therefore not merely extra polish on an already working pipeline: in the pairing where standard OPD stalls, adapting the teacher is what makes it useful.
Configuration (3) sits in between, with OPD gaining 12.39 points over $S_0$ and \spta{} 13.21.

\paragraph{Additional task-level teacher RL does not explain the full gain.}
In configuration (1), Tea-RL-OPD improves over OPD by 0.45 points, while \spta{} improves by 3.78 points, exceeding Tea-RL-OPD by 3.33 points under the matched teacher optimization schedule.
In configuration (2), the corresponding gains over OPD are 2.16 and 3.03 points, yielding a 0.87-point advantage for \spta{}.
Since the two teachers differ only in the state from which their RLVR rollouts start, the extra gain is attributable to prefix conditioning rather than to additional teacher optimization.


\section{Analysis}
\label{sec:analysis}

We analyze teacher recovery in configurations (1) and (2), then examine the second adaptation round in configuration (1).

\subsection{Capability vs.\ Reliability: What Does \spta{} Change?}
\label{sec:capability}

Table~\ref{tab:teacher} separates clean-prompt accuracy from prefix recovery in configurations (1) and (2).
For Qwen3-4B, \spta{} reduces clean accuracy from 60.99\% to 54.39\%, yet improves recovery from 28.91\% to 49.88\% at $r{=}50\%$ and from 22.36\% to 47.68\% at $r{=}90\%$.
The gap between clean accuracy and recovery at $r{=}90\%$ shrinks from 38.63 to 6.71 points.
For DeepSeek-R1-Distill-Qwen-7B, clean accuracy rises by 2.51 points, while recovery improves by 6.28 and 7.32 points at $r{=}50\%$ and $90\%$; its clean-to-90\% gap shrinks from 10.12 to 5.31 points.
Thus the recovery gains exceed the clean-prompt gains in both models, but preservation of clean-prompt performance is model-dependent.
These results do not establish the absence of forgetting on other domains.
\emph{\spta{} targets reliability on student states, not necessarily higher standalone accuracy.}

\begin{table}[t]
\caption{\textbf{Recovery gains are distinct from clean-prompt gains.}
Teacher accuracies are means over five benchmarks (avg@$16$) at each student-prefix ratio $r$; benchmark-level results are in Appendix~\ref{app:recovery}.
$\Delta$ is \spta{} minus Original in percentage points.
The last column gives the corresponding distilled student's average from Table~\ref{tab:main}, not a teacher metric.
The italic reference row in each block is the prefix-source student's own accuracy, which is constant in $r$ (see text) and serves as the break-even level below which the teacher carries no information the student lacks.
}
\label{tab:teacher}
\begin{center}
\small
\setlength{\tabcolsep}{6pt}
\begin{tabular}{lccccccc}
\toprule
& \textbf{Clean} & \multicolumn{4}{c}{\textbf{Recovery}} & \textbf{Student} \\
\cmidrule(lr){2-2}\cmidrule(lr){3-6}\cmidrule(lr){7-7}
\textbf{Teacher} & $r{=}0$ & $r{=}25\%$ & $r{=}50\%$ & $r{=}75\%$ & $r{=}90\%$ & Avg. \\
\midrule
\multicolumn{7}{l}{\textit{(1) Qwen3-4B}} \\
Original       & 60.99 & 32.76 & 28.91 & 24.82 & 22.36 & 31.52 \\
\spta{}  & 54.39 & 50.46 & 49.88 & 48.56 & 47.68 & 35.30 \\
$\Delta$       & $-6.60$ & $+17.70$ & $+20.97$ & $+23.74$ & $+25.32$ & $+3.78$ \\
\addlinespace
\textit{Student $\pi_S$ (reference)} & \textit{12.49} & \textit{12.49} & \textit{12.49} & \textit{12.49} & \textit{12.49} & -- \\
\midrule
\multicolumn{7}{l}{\textit{(2) DeepSeek-R1-Distill-Qwen-7B}} \\
Original       & 46.63 & 40.32 & 37.08 & 37.73 & 36.51 & 30.20 \\
\spta{}  & 49.14 & 44.80 & 43.36 & 44.34 & 43.83 & 33.23 \\
$\Delta$       & $+2.51$ & $+4.48$ & $+6.28$ & $+6.61$ & $+7.32$ & $+3.03$ \\
\addlinespace
\textit{Student $\pi_S$ (reference)} & \textit{29.58} & \textit{29.58} & \textit{29.58} & \textit{29.58} & \textit{29.58} & -- \\
\bottomrule
\end{tabular}
\end{center}
\end{table}

\subsection{Recovery Across Benchmarks and Prefix Depths}
\label{sec:recovery_depth}

The aggregate recovery gains are broadly reflected at the benchmark level (Appendix~\ref{app:recovery}).
Qwen3-4B improves on all 20 benchmark--ratio combinations with nonzero prefixes; DeepSeek-R1-Distill-Qwen-7B improves on 19 of 20.
The exception is HMMT24 at $r=25\%$, where the latter decreases from 25.83 to 25.62.
Thus the positive means are not driven solely by one benchmark, although their magnitudes vary.

The adapted teachers also show smaller endpoint declines as prefixes lengthen from 25\% to 90\%.
For Qwen3-4B, mean recovery decreases by 10.40 points before adaptation (32.76 to 22.36), compared with 2.78 points afterward (50.46 to 47.68).
For DeepSeek-R1-Distill-Qwen-7B, the corresponding declines are 3.81 and 0.97 points.
The adapted teachers are thus markedly less sensitive to how deep into a student trajectory they are queried, which is the regime where OPD spends most of its supervision budget.
These endpoint comparisons describe robustness to prefix depth; they do not identify an optimal truncation ratio, and the trend is not monotonic at every intermediate ratio.

\paragraph{The teacher's advantage over its own student nearly vanishes on student prefixes.}
A natural reference for Eq.~\eqref{eq:rec} is the student itself.
Because the evaluation prefixes are drawn from $\pi_S$, a continuation \emph{by $\pi_S$} from such a prefix is again a sample from $\pi_S(\cdot\mid x)$; the student's self-continuation accuracy therefore equals its clean-prompt accuracy at every ratio $r$, giving the flat reference rows 12.49 and 29.58 in Table~\ref{tab:teacher}.
OPD can only be informative to the extent that the teacher exceeds this level at the states where it is queried.
Measured this way, the original Qwen3-4B teacher leads its student by 48.50 points on clean prompts but by just 9.87 points at $r{=}90\%$, and \spta{} restores the lead to 35.19 points.
For DeepSeek-R1-Distill-Qwen-7B the lead falls from 17.05 to 6.93 points and is restored to 14.25.
This also offers a plausible reading of the near-zero OPD baseline in configuration (2): the teacher that OPD actually queries there is barely stronger than the student it is teaching, so there is little for the student to absorb.
We report this as a consistent pattern across two pairings rather than an established predictive relationship.

Table~\ref{tab:teacher} links these teacher-side findings to stronger distilled students within each pairing.
In particular, Qwen3-4B improves the student despite losing clean-prompt accuracy.
This supports considering recovery alongside standalone accuracy.

\subsection{A Second Round: Refreshing the Student-Prefix Distribution}
\label{sec:iteration_results}

Because distillation moves the student, the prefixes collected before Stage~I gradually cease to describe where the teacher is queried.
We therefore run one additional \spta{} round in configuration (1): the Iter1 student supplies the new prefix source, and both teacher and student are warm-started from the preceding round rather than reset to their original initializations.
We report this round separately because it consumes a second full adaptation and distillation budget and is thus not comparable to the single-round rows of Table~\ref{tab:main}.

Table~\ref{tab:iter2} shows that the second round improves the student average from 35.30 to 36.57, a further $+1.27$ points and a cumulative $+5.05$ points over OPD.
The improvement is consistent rather than concentrated: all five benchmarks gain over Iter1, by 2.08, 0.42, 2.41, 1.25, and 0.21 points.
Two caveats limit what this establishes.
First, the refreshed prefix source is confounded with continued teacher and student optimization, so the numbers reflect the iterative procedure as a whole rather than the isolated effect of re-collecting prefixes.
Second, we do not compare against an equally long OPD run under an unchanged teacher, so the second round supports iteration as a usable extension of \spta{} without demonstrating an accuracy--compute advantage.

\begin{table}[htbp]
\caption{\textbf{Effect of a second \spta{} round} in configuration (1) (Qwen3-4B $\rightarrow$ Qwen3-4B-Base), accuracy (\%, avg@$16$). Iter1 is the \spta{}-OPD row of Table~\ref{tab:main}; Iter2 refreshes the prefix source from the Iter1 student and warm-starts both models from the preceding round, so it uses an additional round of compute and is not matched to the rows above it.}
\label{tab:iter2}
\centering
\small
\setlength{\tabcolsep}{4pt}
\begin{tabular}{lcccccc}
\toprule
\textbf{Method} & \textbf{AIME24} & \textbf{AIME25} & \textbf{AMC23}
  & \textbf{HMMT24} & \textbf{HMMT25} & \textbf{Avg.} \\
\midrule
OPD & 33.54 & 26.04 & 65.51 & 16.04 & 16.46 & 31.52 \\
\spta{}-OPD (Iter1) & 37.71 & 33.54 & 71.91 & 16.04 & 17.29 & 35.30 \\
\addlinespace
\spta{}-OPD-Iter2 & \textbf{39.79} & \textbf{33.96} & \textbf{74.32} & \textbf{17.29} & \textbf{17.50} & \textbf{36.57} \\
$\Delta$ (Iter2 $-$ Iter1) & $+2.08$ & $+0.42$ & $+2.41$ & $+1.25$ & $+0.21$ & $+1.27$ \\
\bottomrule
\end{tabular}
\end{table}

\section{Conclusion}
\label{sec:conclusion}

A teacher's standalone accuracy does not fully describe how useful it is for on-policy distillation.
We introduced \spta{}, which first adapts the teacher with verifiable rewards on continuations of randomly truncated student rollouts and then freezes it for standard OPD, leaving the distillation objective untouched.
A single round improves mean student accuracy in all three pairings and outperforms task-level teacher RL in both pairings that include that control---even for a Qwen3-4B teacher whose clean-prompt accuracy falls while its student improves.
Measured against the student's own self-continuation accuracy, an unadapted teacher's advantage can all but vanish at the very prefixes OPD queries, and this is precisely the pairing in which we observe standard OPD to stall.
Teachers should therefore be prepared and evaluated on the states their students actually visit, with standalone capability and adaptation cost monitored alongside.

\section{Discussion and Limitations}
\label{sec:limitations}

Several questions remain open.
Our experiments show that prefix-conditioned adaptation helps, but not that the benefit is unique to \emph{student}-sourced prefixes, since we evaluate no teacher-sourced or unrelated-model prefixes of comparable length as controls, leaving prefix source and context length partly entangled; teacher continuation accuracy likewise does not establish better token-level calibration or error localization, and although we position \spta{} as complementary to prefix-selection and rollout-allocation methods \citep{earlystoppingopd,pruneopd,prefixguidedopd}, we report no head-to-head or combined comparisons.
Adaptation also carries a model-dependent cost in clean-prompt accuracy---Qwen3-4B loses 6.60 points while DeepSeek-R1-Distill-Qwen-7B gains 2.51---which mixing clean prompts into adaptation or regularizing against the original teacher might mitigate, though we do not test either remedy.
Our scope is mathematical reasoning with verifiable final answers and teachers of 1.5B--7B parameters, so transfer to open-ended tasks, other domains, or much larger teachers remains untested, as does general capability degradation.
Finally, we report avg@$16$ point estimates without multi-seed runs or confidence intervals, so small differences such as the 0.82-point gain in configuration (3) should not be read as established improvements, and we demonstrate neither an accuracy--compute advantage over longer OPD or student RLVR, nor that a teacher adapted to one student transfers to others.

\label{bodyend}
\subsection*{AI use statement}
We used generative AI assistance in two ways.
First, during the early stage of this project, we used it to support literature collection.
Second, we polished the language and corrected grammar in text drafted by the authors.
AI assistance was not used to design the method, run or configure experiments, or produce numerical results; every number reported in this paper comes from the authors' own runs, and no missing measurement was filled in or estimated by an AI system.

\subsection*{Ethics statement}
This work builds entirely on publicly released artifacts.
All teacher and student models (the Qwen3, DeepSeek-R1-Distill, and JustRL families), the training prompts (DAPO-Math-17k), and the evaluation benchmarks (AIME24/25, AMC23, and HMMT24/25) are openly available, and we use them in accordance with their respective licenses and intended research use.
Consequently, the study raises no foreseeable ethical risks and did not require review by an institutional ethics board.


\bibliography{iclr2027_conference}
\bibliographystyle{iclr2027_conference}

\appendix

\section{Implementation Details}
\label{app:impl}

\paragraph{Two-stage configuration.}
Table~\ref{tab:hparam} reports the shared training settings used across all teacher--student pairings.
Stage~I uses GRPO with a group of eight sampled teacher continuations per prefix-conditioned input.
Stage~II uses policy-gradient distillation with four student rollouts per prompt, not task-reward GRPO, even though the trainer configuration retains a GRPO advantage-estimator setting.
The OPD loss disables task rewards and uses only the teacher-derived signal.
Both stages disable reference-KL penalties and entropy regularization; the distillation signal itself is not a reference-KL regularizer.

\begin{table}[htbp]
\caption{Shared training settings across all teacher--student pairings. Input limits apply to the full conditioned input, including the student prefix in Stage~I.}
\label{tab:hparam}
\centering
\small
\setlength{\tabcolsep}{4pt}
\begin{tabular}{lcc}
\toprule
\textbf{Setting} & \textbf{Teacher adaptation} & \textbf{Student OPD} \\
\midrule
Training objective & GRPO & PG-OPD (\texttt{k1}) \\
Inputs per training batch & 256 & 64 \\
Continuations per input & 8 & 4 \\
Sequences per rollout batch & 2048 & 256 \\
PPO mini-batch size (inputs) & 64 & 64 \\
Maximum input length & 7168 & 1024 \\
Maximum generated length & 9216 & 7168 \\
Optimizer & AdamW & AdamW \\
Learning rate & $10^{-6}$ & $10^{-6}$ \\
Weight decay & 0.01 & 0.01 \\
Adam $\beta_1,\beta_2$ & $(0.9,0.999)$ & $(0.9,0.999)$ \\
LR schedule / warmup steps & Constant / 0 & Constant / 0 \\
PPO epochs per batch & 1 & 1 \\
Ratio clipping interval & $[0.8,1.28]$ & $[0.8,1.2]$ \\
Gradient norm clipping & 1.0 & 1.0 \\
Reference-KL / entropy coefficient & 0 / 0 & 0 / 0 \\
Rollout temperature & 1.0 & 1.0 \\
Training epochs per round & 5 & 1 \\
Validation interval & 10 steps & 10 steps \\
Separate teacher-inference GPUs & -- & $1\times8$ \\
Rollout tensor parallelism & 1 & 1 \\
\bottomrule
\end{tabular}
\end{table}

\paragraph{Optimization and execution.}
Training uses FSDP, dynamic batching, padding removal, and gradient checkpointing, with vLLM for generation and teacher inference.
The framework uses BF16 mixed precision, with FP32 reductions by default.
Actor parameters and optimizer states are not offloaded.
Overlength inputs are filtered rather than silently truncated; the configured input-plus-generation budgets are 16,384 tokens for teacher adaptation and 8192 for student OPD.

\paragraph{Policy-gradient distillation signal.}
Equations~\eqref{eq:opd} and~\eqref{eq:spta} give the distribution-level formulation.
The implementation uses sampled-token reverse-KL feedback rather than an exact full-vocabulary KL sum.
For a sampled token, write $s_t=\log\pi_\theta(y_t^S\mid x,y^S_{<t})$ and $q_t=\log\pi_{\phi^\star}(y_t^S\mid x,y^S_{<t})$.
The \texttt{k1} estimate and policy-gradient signal are
\begin{equation}
 d_t=s_t-q_t,\qquad A_t=-\operatorname{stopgrad}\!\left(\operatorname{clip}(d_t,-10,10)\right).
\label{eq:pgopd_signal}
\end{equation}
This is a local token signal, not a discounted reward-to-go or a group-normalized task advantage.
The policy loss uses a PPO-style clipped surrogate with ratio interval $[0.8,1.2]$ and negative-advantage dual clipping with coefficient 3; its configured aggregation is response-token mean.
Although the launch configuration also sets \texttt{log\_prob\_min\_clamp} to $-10$, the \texttt{k1} implementation does not apply that parameter; the active clamp is on $d_t$.

\paragraph{Training schedule and validation.}
Each round uses a five-epoch teacher-adaptation schedule and a one-epoch student-distillation schedule.
In Iter2, both models continue from the preceding round rather than resetting to the original initialization.
Both launch configurations validate on AIME25 with temperature 0.7, top-$p$ 0.95, and 16 sampled responses, every ten training steps.

\paragraph{Data provenance.}
The OPD training file is \path{dapo-math-17k}.
Three counts should not be conflated: the $m{=}4$ student rollouts generated per prompt, the single rollout among them that is retained and truncated into a prefix, and the eight teacher continuations GRPO samples per prefix during Stage~I.

\paragraph{Recovery protocol.}
The prefix source is the initial student for each pairing.
Each benchmark problem contributes one student rollout, from which we retain $k=\lfloor rL\rfloor$ tokens for $r\in\{0,0.25,0.50,0.75,0.90\}$; the original and Iter1 teachers are then compared on exactly this prefix.
Because $L$ varies across problems, a fixed ratio corresponds to a different absolute context length for each problem: the ratio specifies a relative position in the student's reasoning, not a token budget or an error location.
Each prefix receives 16 sampled teacher continuations, matching the 16 samples used for student evaluation, so all reported accuracies in this paper are avg@$16$; this is independent of the eight continuations per prefix used for GRPO during Stage~I training.
A correct final answer receives score 1 and an incorrect answer receives score 0; scores are averaged over the 16 continuations and then over problems, rather than aggregated by best-of-$n$ selection.

\paragraph{Interpreting the reported averages.}
Every benchmark receives equal weight in Avg., irrespective of its number of problems.
Tables retain scores rounded to two decimal places, so re-averaging displayed benchmark scores can differ from the reported mean by 0.01 points.
The 16 response samples are evaluation draws, not 16 independently trained models.

\section{Full Teacher Recovery Results}
\label{app:recovery}

Table~\ref{tab:recovery_full} reports the benchmark-level results underlying Figure~\ref{fig:motivating} and Table~\ref{tab:teacher}.
For each teacher--student pairing, Original and \spta{}-Iter1 use identical sampled student prefixes at each ratio.
Prefixes come from Qwen3-4B-Base in configuration (1) and DeepSeek-R1-Distill-Qwen-1.5B in configuration (2).
Ratios are fractions of each student's generated token sequence, not fixed token budgets.
At $r{=}0$, the teacher solves the problem without a student prefix.
The Avg.\ column is the unweighted mean across the five benchmarks, with reported values rounded to two decimals.

\begin{table}[htbp]
\caption{\textbf{Full teacher recovery results.} Accuracy (\%, avg@$16$) at each student-prefix ratio. All adapted teachers are from Iter1. These are teacher continuation results, not distilled-student scores.}
\label{tab:recovery_full}
\centering
\small
\setlength{\tabcolsep}{4pt}
\begin{tabular}{llcccccc}
\toprule
\textbf{Teacher} & \textbf{Prefix} & \textbf{AIME24} & \textbf{AIME25} & \textbf{AMC23} & \textbf{HMMT24} & \textbf{HMMT25} & \textbf{Avg.} \\
\midrule
\multicolumn{8}{l}{\textit{(1) Qwen3-4B}} \\
Original & 0\%  & 72.08 & 64.17 & 88.93 & 35.42 & 44.37 & 60.99 \\
Original & 25\% & 38.33 & 27.08 & 64.01 & 17.71 & 16.67 & 32.76 \\
Original & 50\% & 31.67 & 23.54 & 63.10 & 15.00 & 11.25 & 28.91 \\
Original & 75\% & 26.25 & 21.67 & 57.23 & 11.87 & 7.08  & 24.82 \\
Original & 90\% & 25.00 & 16.88 & 53.24 & 10.21 & 6.46  & 22.36 \\
\addlinespace
\spta{}-Iter1 & 0\%  & 59.58 & 58.13 & 88.40 & 32.71 & 33.12 & 54.39 \\
\spta{}-Iter1 & 25\% & 55.00 & 50.21 & 85.84 & 29.58 & 31.67 & 50.46 \\
\spta{}-Iter1 & 50\% & 56.46 & 49.79 & 85.24 & 28.12 & 29.79 & 49.88 \\
\spta{}-Iter1 & 75\% & 53.54 & 49.38 & 84.49 & 26.87 & 28.54 & 48.56 \\
\spta{}-Iter1 & 90\% & 54.17 & 46.25 & 84.26 & 26.87 & 26.87 & 47.68 \\
\midrule
\multicolumn{8}{l}{\textit{(2) DeepSeek-R1-Distill-Qwen-7B}} \\
Original & 0\%  & 56.04 & 38.96 & 82.76 & 28.54 & 26.87 & 46.63 \\
Original & 25\% & 45.00 & 32.29 & 77.03 & 25.83 & 21.46 & 40.32 \\
Original & 50\% & 42.08 & 27.29 & 74.17 & 23.13 & 18.75 & 37.08 \\
Original & 75\% & 43.13 & 30.00 & 74.70 & 22.08 & 18.75 & 37.73 \\
Original & 90\% & 42.71 & 28.96 & 72.97 & 21.67 & 16.25 & 36.51 \\
\addlinespace
\spta{}-Iter1 & 0\%  & 57.71 & 50.62 & 80.72 & 33.33 & 23.33 & 49.14 \\
\spta{}-Iter1 & 25\% & 55.83 & 39.37 & 80.87 & 25.62 & 22.29 & 44.80 \\
\spta{}-Iter1 & 50\% & 54.58 & 37.92 & 80.35 & 24.17 & 19.79 & 43.36 \\
\spta{}-Iter1 & 75\% & 54.79 & 36.88 & 79.82 & 27.71 & 22.50 & 44.34 \\
\spta{}-Iter1 & 90\% & 53.12 & 37.50 & 80.42 & 25.83 & 22.29 & 43.83 \\
\bottomrule
\end{tabular}
\end{table}

\end{document}